\documentclass{article}

\usepackage{naacl2021}

\usepackage[utf8]{inputenc} 
\usepackage[T1]{fontenc}    
\usepackage{hyperref}       
\usepackage{url}            
\usepackage{booktabs}       
\usepackage{amsfonts}       
\usepackage{nicefrac}       
\usepackage{microtype}      
\usepackage{lipsum}
\usepackage{graphicx}

\graphicspath{ {./images/} }
\usepackage{fancyhdr}
\usepackage[symbol]{footmisc}
\usepackage{caption} 
\usepackage{subcaption}
\usepackage{float}
\usepackage{adjustbox}
\usepackage{amsmath}
\usepackage{multirow}
\usepackage{lscape}
\usepackage{longtable}

\usepackage{multiaudience}

\SetNewAudience{long}

\definecolor{mygrey}{rgb}{0.758, 0.758, 0.758}

\newcommand\obrazek[1]{
  (Figure \ref{#1})}

\title{Czech News Dataset for Semantic Textual Similarity }

\author{
  Jakub Sido \textsuperscript{1,2} 
  \and
  Michal Seják \textsuperscript{1}
  \and
  Ondřej Pražák \textsuperscript{1,2}\\ 
  \and
  \bf{Miloslav Konopík} \textsuperscript{1,2}
 \and
 Václav Moravec \textsuperscript{3} \\[0.5em]
\tt{\{sidoj,sejakm,ondfa,konopik\}@kiv.zcu.cz}\\[0.5em]
 \textsuperscript{1} NTIS -- New Technologies for the Information Society\\
  \textsuperscript{2}Department of Computer Science and Engineering,\\
Faculty of Applied Sciences, University of West Bohemia, Czech Republic\\
  \textsuperscript{3}Department of Journalism,
  Faculty of Social Sciences, Charles University, Czech Republic
  \\
}

\begin{document}
\maketitle
\begin{abstract}

This paper describes a novel dataset consisting of sentences with two different semantic similarity annotations; with and without surrounding context. The data originate from the journalistic domain in the Czech language. 
The final dataset contains 138,556 human annotations divided into train and test sets.
In total, 485 journalism students participated in the creation process. To increase the reliability of the test set, we compute the final annotations as an average of 9 individual annotation scores. We evaluate the dataset quality measuring inter and intra annotator agreements. Besides agreement numbers, we provide detailed statistics of the collected dataset. We conclude our paper with a baseline experiment of building a system for predicting the semantic similarity of sentences. Due to the massive number of training annotations (116,956), the model significantly outperforms an average annotator (0.92 versus 0.86 of Pearson's correlation coefficient).






\end{abstract}


\section{Introduction}

This paper describes a novel dataset consisting of sentences with semantic similarity annotations. The dataset comprises pairs of sentences in the Czech language, where each pair is associated with a similarity score. The purpose of the dataset is to train and evaluate systems for predicting the semantic similarity of sentences.

Currently, the NLP field relies ever more on unsupervised or self-supervised models. Nevertheless, well-annotated datasets are still required for model adaptation or testing. We pay greater attention to the testing part of the dataset. For this part, every sentence pair is annotated independently by nine annotators. Consequently, we store the average of the independent annotations as the final score. For the training part, we prefer to cover as much diverse data as possible. Therefore, we stick with one annotation per sentence pair.

Current NLP models\footnote{For example, the Longformer model \cite{beltagy2020longformer}.} are becoming increasingly capable of processing sentences in their contexts. Therefore, we include the context of sentences when creating the training pairs. Finally, we annotate the similarity scores with and without the context.

We decided to cooperate with journalism students to produce better annotations, since they are generally skilled at handling text data. To improve the annotations further, professional journalists supervised the student annotators. We believe that the contribution of skilled annotators increases the quality of the dataset. 

The final dataset contains 138,556 human annotations. Its creation required a considerable amount of human annotation work -- 485 annotators and the time spent creating the dataset was around 1,017 man-hours.






\section{Related Work}

Regarding English, there are many datasets for semantic textual similarity (STS). The most commonly used datasets come from the SemEval competition. There have been six competitions on STS since 2012. \cite{agirre2012semeval, agirre2013sem, agirre2014semeval, agirre2015semeval,agirre2016semeval,cer2017semeval}. These datasets include pairs of sentences taken from news articles, forum discussions, headlines, image and video descriptions labeled with a similarity score between 0 and 5. The goal is to evaluate how the cosine distance between two sentences correlates with a human-labeled similarity score through Pearson and Spearman correlations. The datasets from all SemEval STS competitions are gathered in the SentEval corpus \cite{conneau2018senteval}. 

As for Czech, an STS dataset exists.  \citet{svoboda2018czech} created this dataset by translating English STS from SemEval to Czech. Then they annotated the sentences they had translated. The main drawback of this dataset is its small size (1425 sentence pairs) in comparison to the dataset presented in this paper. 

\section{Source Data} \label{sec:dataset}
The raw text data for the introduced dataset come from the Czech News Agency (CNA).
CNA delivers complete service for Czech journalists, including images, quick short news, reports, observing long-term incidents across many domains. CNA staff publish news via an internal standardized process. Each event CNA follows is internally called an  \textit{incident}. Every incident is associated with partial \textit{news}, which update the incident over time (we call this news \textit{reports} hereinafter in this paper). The incidents are (usually) concluded by a human made \textit{summary}. We employ summaries and reports grouped by incidents in our dataset. 


The original database of summaries and reports is a private property of CNA. To publish a part of the original data, we agreed to construct the dataset so that it prevents the full reconstruction of the original data.We only publish isolated sentences from summaries and reports with limited contexts. We remove all the relations between reports. 


\section{Process of Collecting Humans Annotations}
The text data for annotation come from our collaboration with CNA. We attempt to help journalists to automatically analyze related reports when creating summaries. Our goal is to train a model for assembling summaries from original reports -- or at least, to help the journalists write summaries. To obtain relevant data for model training, we need to pair sentences from reports and summaries. 

\begin{figure}[t]
     \centering
      \begin{subfigure}[b]{0.45\textwidth}
         \centering
         \includegraphics[width=\textwidth]{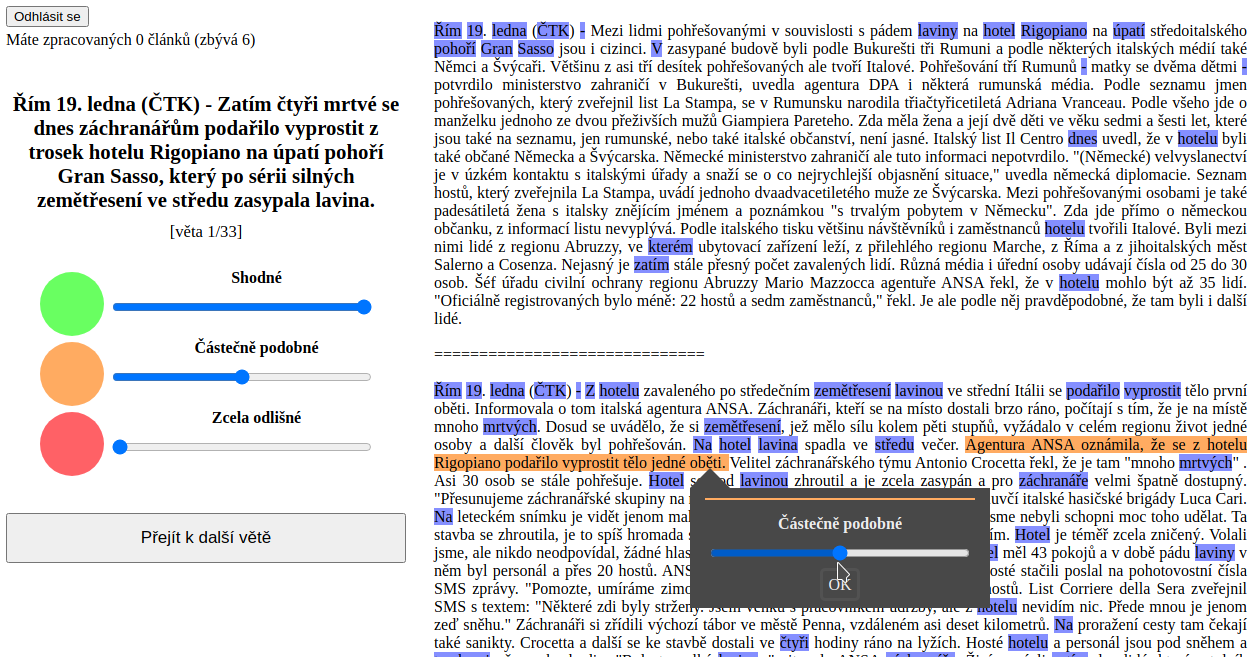}
         \caption{The screen of the first round.}
         \label{fig:ann_a}
     \end{subfigure}
    \\[1.5em]
     \begin{subfigure}[b]{0.45\textwidth}
         \centering
         \includegraphics[width=\textwidth]{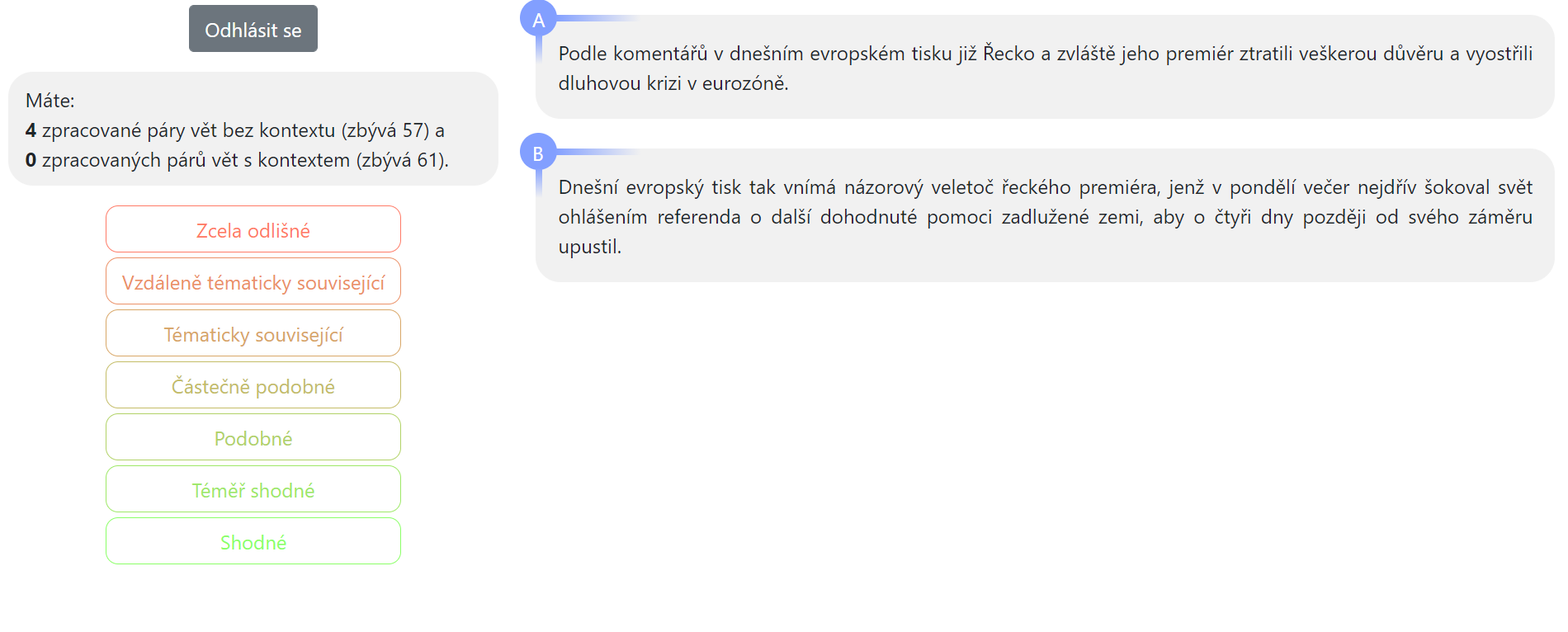}
         \caption{The screen of context free annotation phase -- second round}
         \label{fig:ann_b}
     \end{subfigure}
     \\[1.5em]
     \begin{subfigure}[b]{0.45\textwidth}
         \centering
         \includegraphics[width=\textwidth]{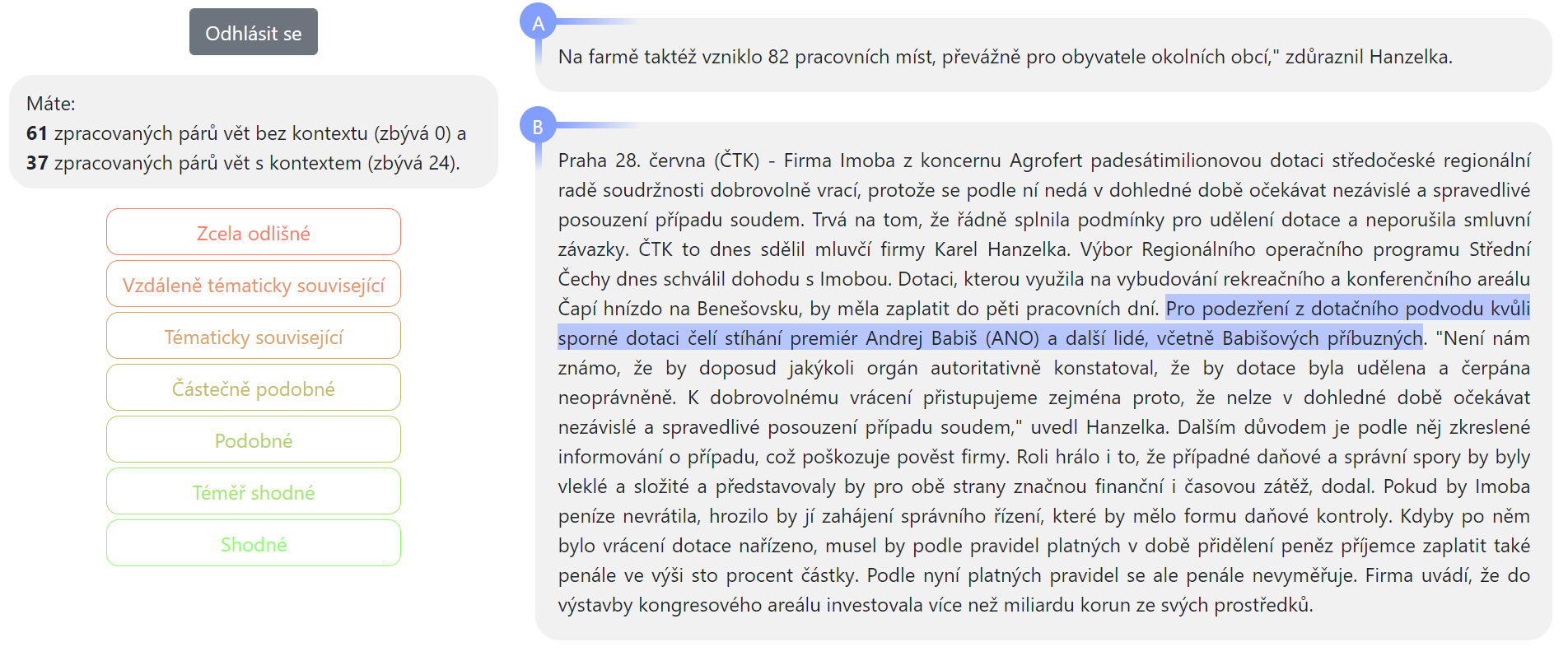}
         \caption{The screen of context dependent annotation phase -- second round}
         \label{fig:ann_c}
     \end{subfigure}
    \caption{Annotation application window screens.}
    \label{fig:ann}
\end{figure}

We have built a web application to collect human annotations of sentence pair similarities to aid the annotation process.  The first sentence of the pair belongs to the summary, and the other one belongs to the original reports (See Figure \ref{fig:ann}). The annotators were asked to give two elementary pieces of information: 
\begin{enumerate}
    \item context-free semantic textual similarity,
    \item context-dependent semantic textual similarity.
\end{enumerate}

The sentences are taken from reports and their summaries; therefore, we can reasonably expect related and semantically similar sentences to be present.


\paragraph{Human Resources}

In this work, we cooperated with two groups of journalist students. Group 1 (272 people) annotated the data in the \textit{first round}. They picked sentence pairs and annotated similarities -- these data are used for the training part of the dataset. 
Group 2 (229 people) participated in the \textit{second round} to create the test part of the dataset. We discuss the process in the following sections in more detail.



\subsection{First Round -- R1}


In the first round (R1), we asked the annotators to choose three sentences ($A_{n}, B_{n}, C_{n}$) from reports for each sentence ($S_n$) in the summary . We instructed them to select the most similar sentence ($A$), the least similar sentence ($C$), and something in the middle ($B$), in order to create a more balanced dataset \obrazek{fig:r1}. To aid the annotators, we highlighted the words from summary sentences in all reports. (Figure \ref{fig:ann_a}). 
We asked the users to use a slide bar to annotate the degree of similarity on a scale of 0--6. The degrees were labeled to give the annotators more easily interpretable possibilities. The labels are: 0: \textit{Completely different}, 1: \textit{Somewhat thematically related}, 2: \textit{Thematically related}, 3: \textit{Partially similar}, 4: \textit{Similar}, 5: \textit{Almost identical}, 6: \textit{Identical}.

\begin{figure*}[ht!]
         \centering
      \includegraphics[width=0.9\textwidth]{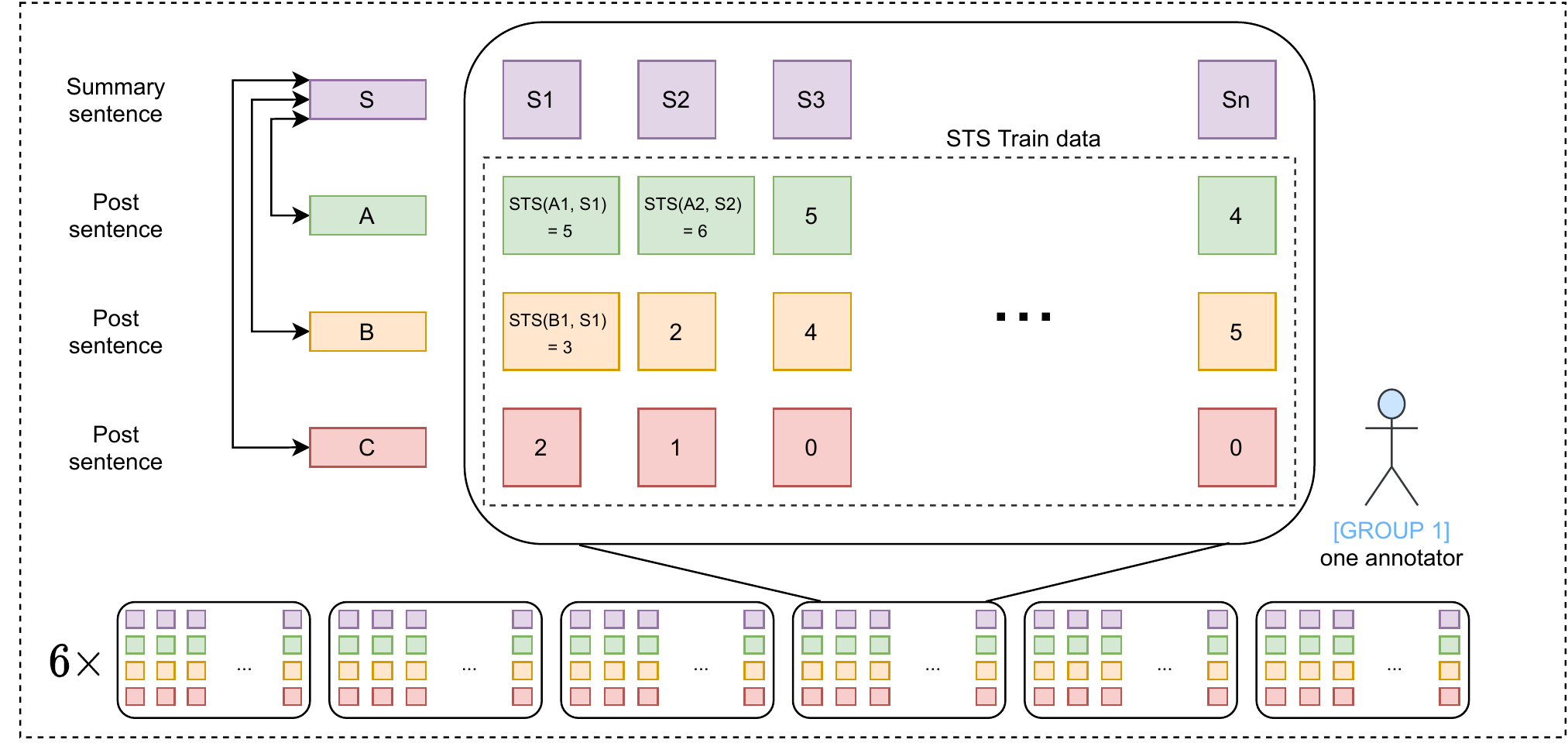}
    \caption{R1 -- The first round of annotation. \textit{S} = summary sentence; \textit{A,B,C} = report sentences. Pairs \textit{SA}, \textit{SB} and \textit{SC} were selected and annotated by Group 1. One annotator has processed 6 full incidents (no overlap). The STS scores for each pair unused in the second round are -- together with the corresponding sentences -- labeled as the \textit{training dataset}.}
    \label{fig:r1}
\end{figure*}

\subsection{Second Round -- R2}
\begin{figure*}[ht!]
    \centering
    \includegraphics[width=\textwidth]{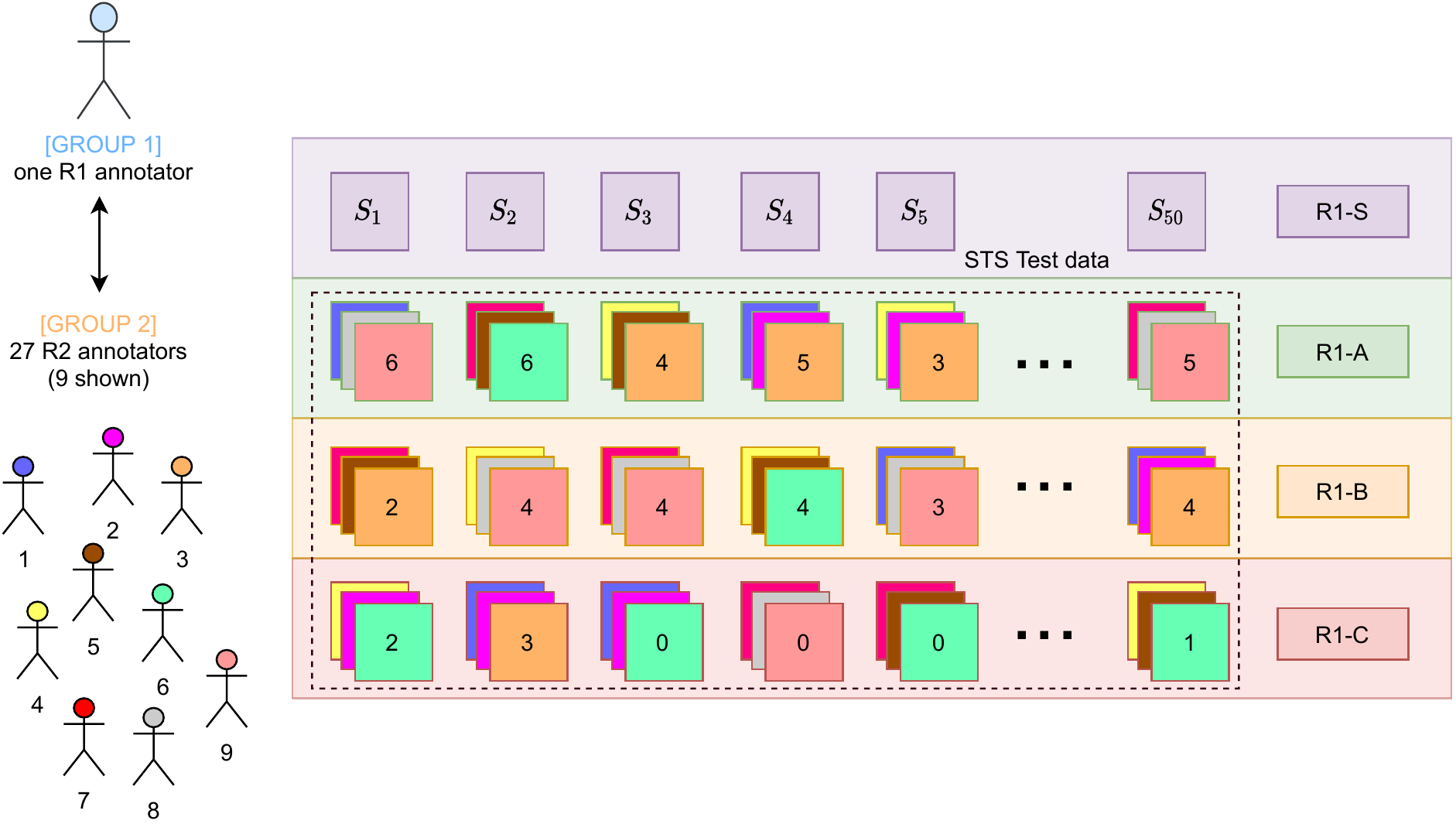}
    \caption{
    R2 -- $S_n$= summary sentence from R1; $A_n, B_n, C_n$ = report sentences from R1. The distinct sequence for each validator.} 
    \label{fig:r2}
\end{figure*}

We created the testing part of the dataset in the second round (R2). We asked the annotators to assign similarity numbers to the sentence pairs (SA, SB, SC) created in R1 on the same scale (0--6). 
The pairs were picked randomly but not independently of one another, as they were specifically crafted  for their union to yield the full original \textit{(S, A, B, C)} tuple set while having an empty intersection (See Figure \ref{fig:r2}). Annotators processed the individual sequences in R2 in the same order as in R1.


\begin{figure*}[ht!]
     \centering
     \begin{subfigure}[t]{0.48\textwidth}
         \centering
         \includegraphics[width=\textwidth]{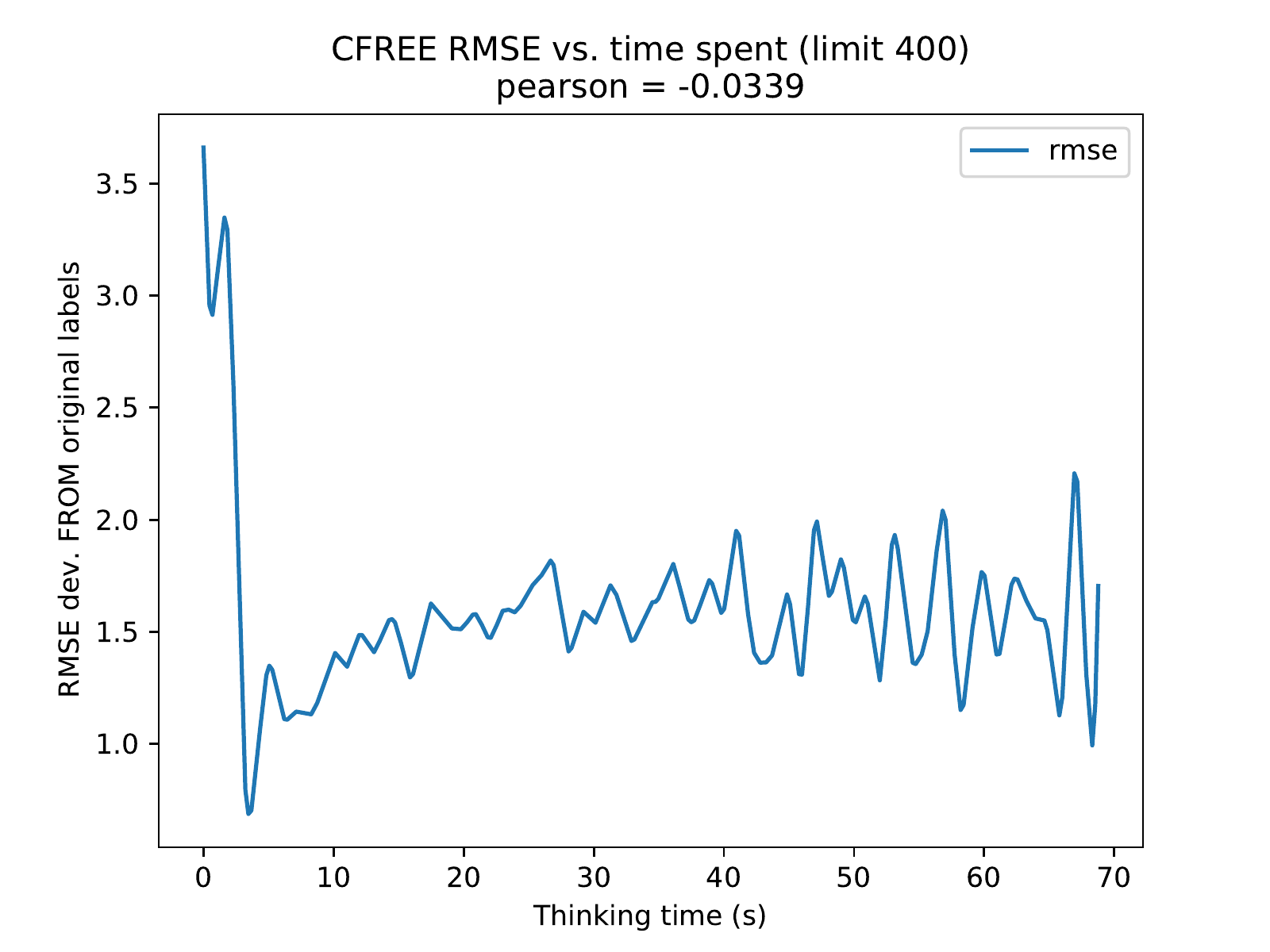}
         \caption{The dependence of root mean squared error (= RMSE) between the R1 score and R2 score on estimated validator (R2) thinking time. All annotations were grouped into buckets (bucket size = 1 second) and RMSE deviation from the original R1 score was calculated for each group. The initial spike corresponds to users who have not given the answer any significant thought.}
         \label{}
    \label{fig:sts-rmse-time}
     \end{subfigure}
     \hfill
     \centering
     \begin{subfigure}[t]{0.48\textwidth}
         \centering
         \includegraphics[width=\textwidth]{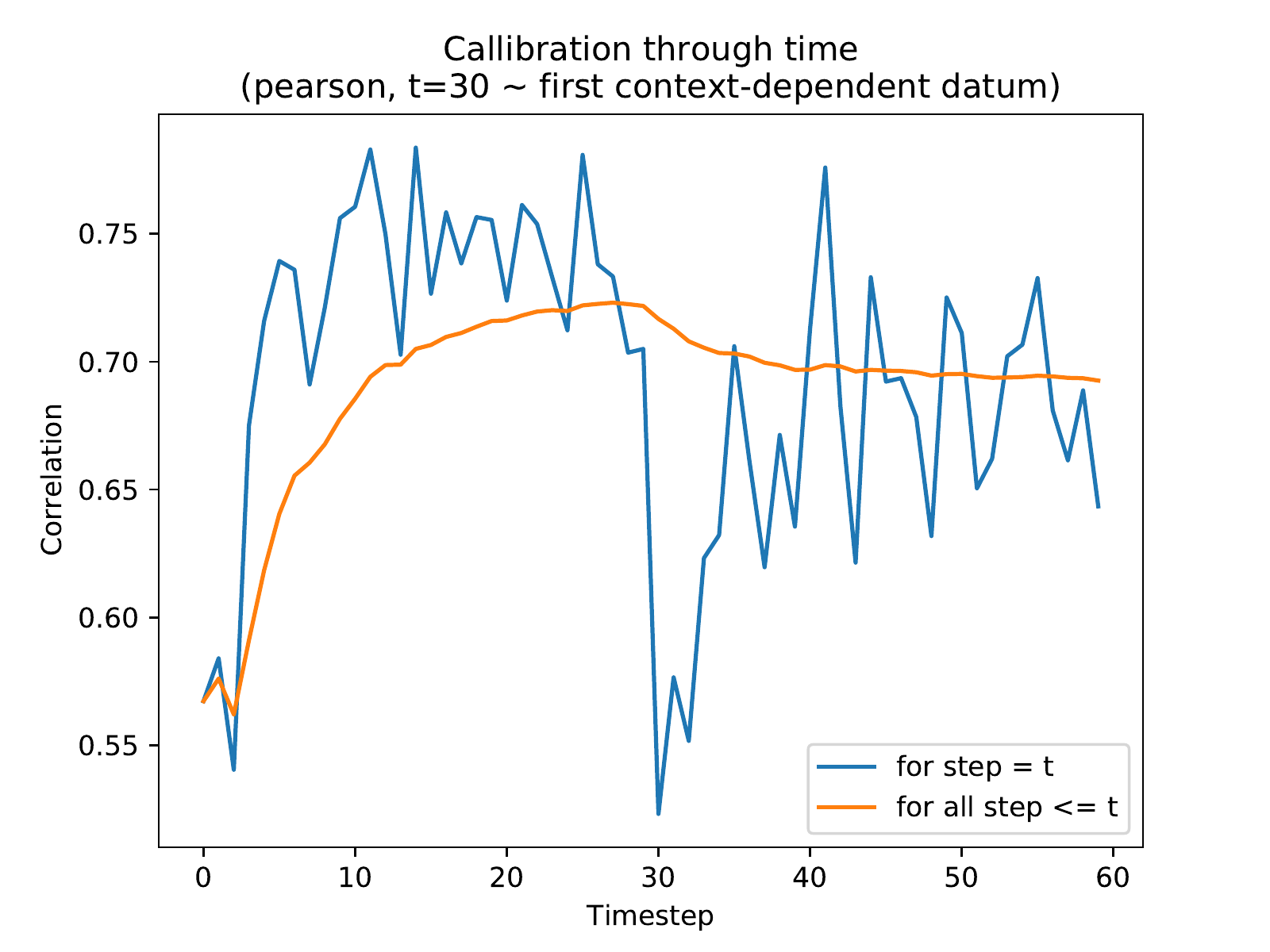}
         \caption{The dependence of pearson correlation between STS scores of R1 and R2 on the number of annotations already made by individual R2 annotator (= calibration curve). Note the value shown is not an average correlation between each R1 annotator and the corresponding R2 annotator, but the correlation between all R1 annotators and all R2 annotators at once.}
         \label{fig:r2_call_a}
     \end{subfigure}
     
     \hfill
     \caption{Statistics of the short preliminary annotation phase. As the true time between submitting annotations consists of thinking time and waiting time (precisely, their sum, which was the only information known to us), the thinking, respective waiting times shown here were estimated by ignoring the other part. 
     }
\end{figure*}

\label{sec:exploratory-r2}



\label{sec:validation-r3}



About two months before the main annotations, we run a short preliminary annotation phase to collect some data for future intra-annotator agreement investigation.

We added a strict rule into the user interface, which forces a 72-hour pause between context-free and context-dependent annotations. 
On the end of both, context-free and context-dependent, we added special blocks to measure intra-annotator agreement.

\begin{figure*}[ht!]
    \centering
    \includegraphics[scale=0.75]{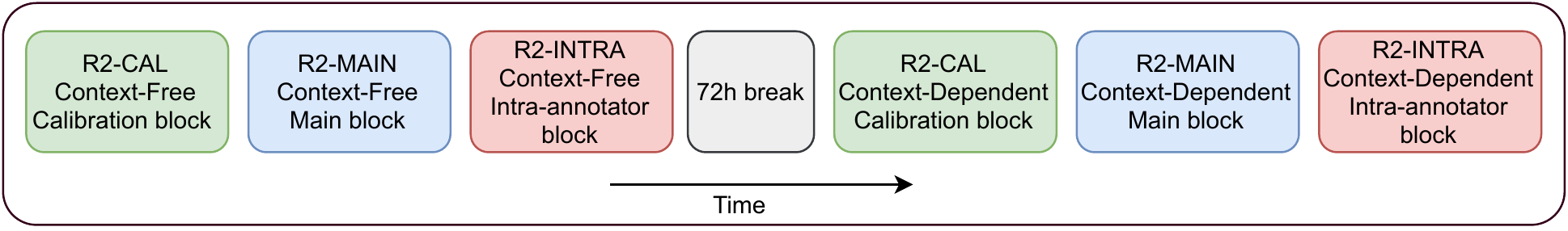}
    \caption{The full round two (R2) on a temporal axis. The target sentence pairs in context-free and context-dependent subphases of each phase were equal and ordered the same exact way. The 72-hour break between the context-free and context-dependent halves has been forced.} 
    \label{fig:r3_overal}
\end{figure*}

For an overview of the second round see Figure \ref{fig:r3_overal}.

\paragraph{Calibration Block (R2-CAL)}
\label{par:calibration-block}
During the short preliminary annotation phase, we noticed a decreased inter-annotator agreement at the begging of annotators' work (see Figure \ref{fig:r2_call_a}). We estimate the number of leading pairs required for successful user calibration to be 6. Therefore, we add six calibration sentence pairs before the main annotation phase. The pairs were manually chosen so that their STS scores cover the full annotation scale. 

\paragraph{Main R2 block (R2-MAIN)}
In the main R2 block, the annotators worked with the pairs that compose the testing part of the dataset. For the better precision of the annotations, we decided to use nine annotations for one sentence pair to get rid of the noise. The samples were randomly shuffled to eliminate any potential STS bias resulting from repeating a stack of annotators with similar semantic intuition. 

We chose the main block size to be 50 to match the expected spent time in man-hours. Consequently, every single individual annotator annotated precisely 50 STS pairs, first without context and then including context.

\paragraph{Intra-annotator block (R2-INTRA)}

\begin{figure*}[h!]
    \centering
    \includegraphics[scale=0.9]{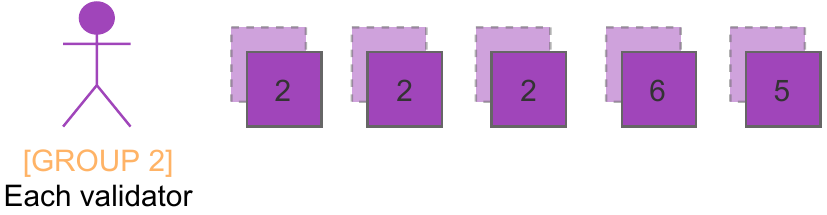}
    \caption{R2-INTRA -- Intra-annotator block -- Five already annotated pairs is randomly sampled and presented to the same annotator to get intra-annotator agreement for both versions -- context-free and context dependent.}
    \label{fig:r3ia}
    
\end{figure*}

After the main block, each annotator was asked to annotate five extra sentence pairs\footnote{Five for context-free and five for context dependent -- ten pairs in total.}, which they already annotated in the short preliminary annotation phase (See Figure \ref{fig:r3ia}). This data was used to measure the intra-annotator agreement.

\section{Dataset Statistics}
The final dataset is divided into the train and test parts. The train part contains 116 956 samples.
In the test dataset, we decided to annotate 1200 pairs for both context free and dependent variants. We designed the process to end up with nine annotations for each pair. However, against our effort, the context-dependent dataset is about 6 \% smaller. However, only 21 out of 1200 sentences have less than seven annotations.

We estimate the time spent creating the training dataset (R1) 876 man-hours (269 annotators, 3.26 hours each on average). Annotators created in the test set in  141 man-hours (216 annotators, around 40 minutes each on average). 

\paragraph{Annotator Agreement} 
To discover the limits of human agreement, we computed a few of metrics suitable for this purpose. 

We store ten annotations made by users in a~short preliminary annotation phase, which we let the users annotate again later. Since each user annotated ten pairs (five for both context-free and context-dependent), we can evaluate intra-annotator agreement. Scores produced by all users in both context phases are concatenated into two vectors (preserving their order). We compute the Pearson and Spearman correlation between these two vectors to quantify how correlated previous scores are to the scores of the same pairs and same users annotated later. We employ the correlation coefficients together with the Mean Squared Error (MSE) and Root Mean Squared Error (RMSE) scores to show how much the annotations of individual annotators differ on average.

For inter-annotator agreement, we took inspiration from \cite{agirre2014semeval}, where the authors compute the Pearson correlations between annotations of each user and the mean of annotations of the other users (on the corresponding sentence pairs) and then average the individual correlations. Apart from Pearson correlations, however, we have decided to compute the average Spearman correlation, average RMSE, and average MSE; again, to assess the scale on which these scores differ.  We present the agreement metrics in Table \ref{tab:agreement}.

Imagine now that we possess an oracle for STS, a theoretical machine that always returns the true STS for any pair of sentences. It is helpful to analyze its performance on our dataset, since it theoretically provides a bound for the quality of any machine learning model evaluated on our dataset. Of course, our dataset does not contain true STS values; human annotation processes are generally noisy. Since we are unable to find a theoretical bound for correlations, we calculate and compute a~lower bound for MSE only.

\begin{table}[]
\centering
\begin{tabular}{ccccc}
\toprule
Agreement         & pcorr & spcorr & RMSE & MSE\\
                \midrule
inter       & 0.832  & 0.777  & 1.140 & 1.461  \\
intra        & 0.746  &  0.719 & 1.396 & 1.948  \\
\bottomrule
\end{tabular}

\caption{Intra-annotator and inter-annotator agreement measurement results. \textit{pcorr} and \textit{spcorr} stand for Pearson and Spearman correlations respectively.}
\label{tab:agreement}
\end{table}

\paragraph{Theoretical lower bound for MSE}

Within the context of a single sentence pair, assuming that the STS scores supplied by our annotators in R2 are approximately normally distributed, it follows that the random variable $\frac{\bar{x} - \mu}{\sigma / \sqrt{n}}$ (where $\bar{x}$ is the STS sample mean, $\mu$ is the true STS mean, $\sigma$~is the STS sample standard deviation and $n$ is the amount of averaged STS results) has the Student's t-distribution with 8 degrees of freedom (since we have 9 annotators per one example). Scaling this random variable by a factor of $\frac{\sigma}{\sqrt{n}}$, we obtain the distribution of $\bar{x} - \mu$, whose variance is the MSE between an oracle for STS (which always returns the true mean $\mu$) and the corresponding STS mean $\bar{x}$, which is in our dataset. Since $\text{Var}[a X] = a^2 \text{Var}[X]$ and the variance of Student's distribution with 8 degrees of freedom is $1.\bar{3}$, we estimate the lower bound for MSE as the average of $1.\bar{3} \frac{\sigma^2}{n}$ for all sentence pairs, which is approximately $0.1731$.

				



\begin{figure*}[ht!]
    \centering
    \begin{subfigure}[t]{0.48\textwidth}
        \centering
        \includegraphics[width=\textwidth]{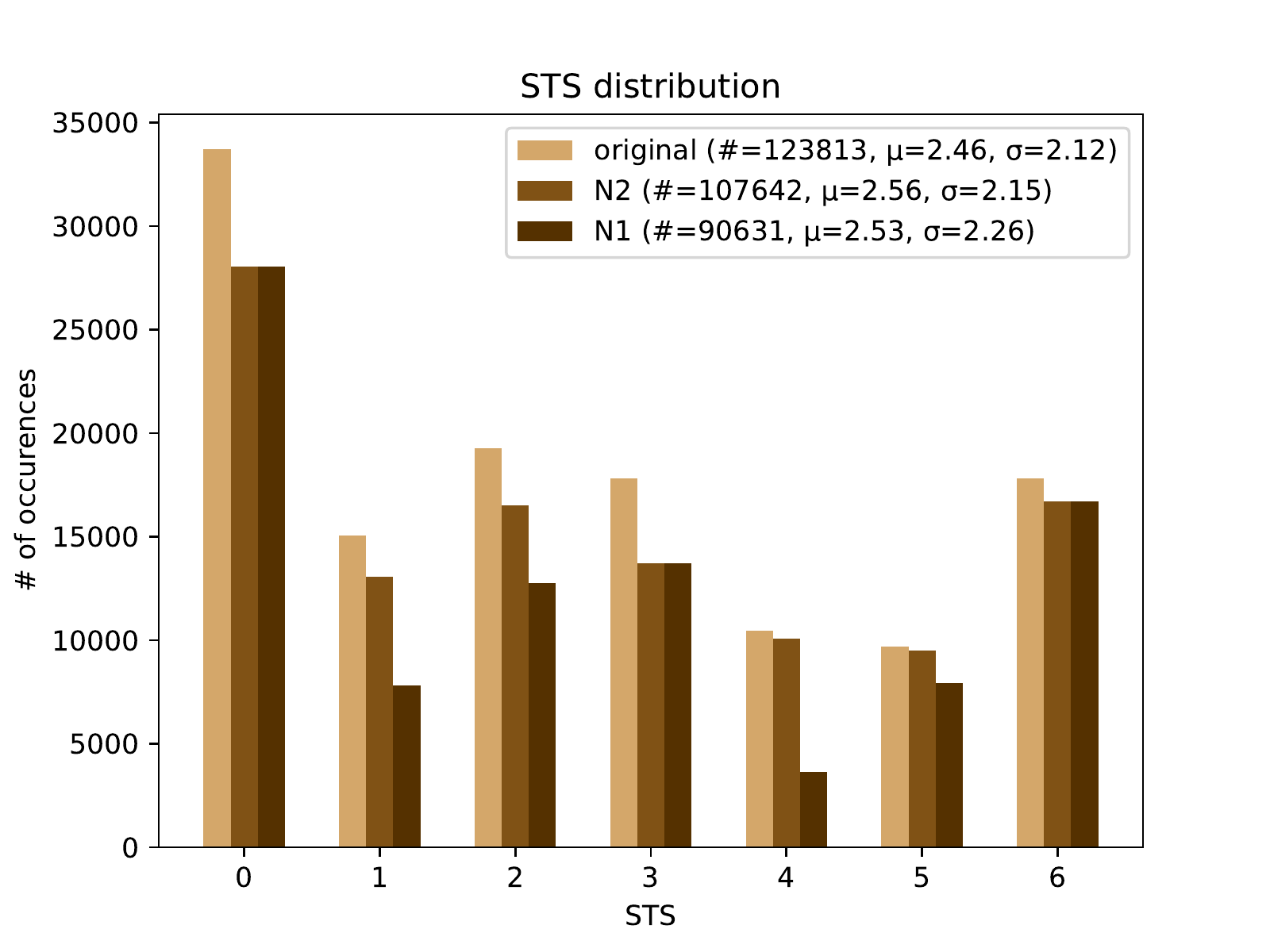}
        \captionsetup{justification=centering}
        \caption{STS distribution in the train dataset before\\ and after performing the individual diagonal filters.}
    \end{subfigure}
    \hfill
    \begin{subfigure}[t]{0.48\textwidth}
        \centering
        \includegraphics[width=\textwidth]{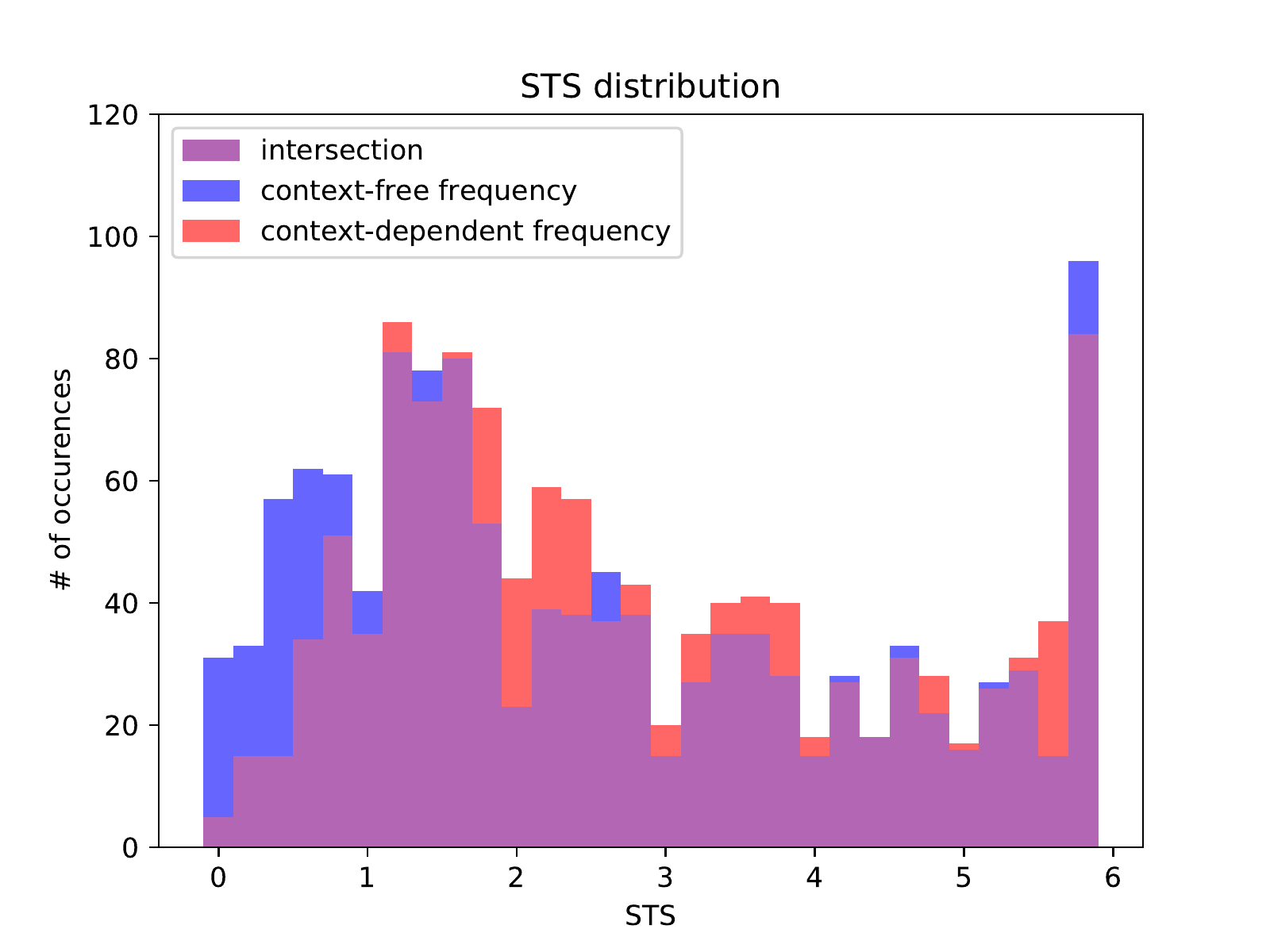}
        \captionsetup{justification=centering}
        \caption{STS distribution in the test dataset. Due to the significantly smaller granularity of test STS (means of 9), the data has been rounded down to units of size 0.2 on the STS scale for clarity.}
    \end{subfigure}
    
    \caption{Train/test dataset STS distributions. 
    }
    \label{fig:sts_distributions}
\end{figure*}

\begin{table*}[]
\centering
\begin{tabular}{lccccc}
\toprule
\textbf{R1 data} & \textbf{pw R1/R2 cfree} & \textbf{pw R1/R2 cdep} &\textbf{inters. R1 and R2}\\ \midrule
\textbf{RAW}             & 0.8787      & 0.8733 &1200 (100 \%)        \\ 
\textbf{N2}        & 0.8831      & 0.8792 & 1101 (92 \%)      \\ 
\textbf{N2-rand}        & 0.8782      & 0.8736 & 1101 (92 \%)      \\ 
\textbf{N1}        & 0.8932      & 0.8918 & 913 (76 \%)         \\ 
\textbf{N1-rand}        & 0.8759      & 0.8714 & 913 (76 \%)         \\ 
\bottomrule
\end{tabular}
\caption{Correlations between STS scores of different R1 filtering methods and the resulting test dataset.  The filtering method indirectly removes sentence pairs from the test dataset as well when computing correlation; the size of the remainder, which was used to calculate correlation with, is shown in the rightmost column. The imaginary set of R1 data created not by our filtering method, but by filtering random elements so as to preserve the original test size, is labelled \textit{-rand}, and the "filtering method" is not actually applied, only the intersection size is preserved. \textit{RAW} is simply the whole of R1 without filtering.}
\label{tab:corr}
\end{table*}

\paragraph{Correlation between original STS scores and test datasets}
For completeness, we present the correlations of STS scores between the scores captured in R1 and the corresponding test context-free and context-dependent datasets (see the first row in Table \ref{tab:corr}). We can observe that both of the test datasets correlate strongly with the original R1 scores. 
Notably, the correlation between context-free and context-dependent STS scores collected in R2 (and compiled into the test datasets) is very high. 
The results of our analysis of equivalence of context-free and context-dependent STS scores (visualised in suggest that only 7.5 \% of the test dataset (90~pairs) are significantly different from their contextual counterpart (See Section \ref{sec:Semantic_shift}).

\paragraph{Influence of context on STS}

\begin{figure}[ht!]
    \centering
    \begin{subfigure}[b]{0.45\textwidth}
        \centering
        \includegraphics[width=\textwidth]{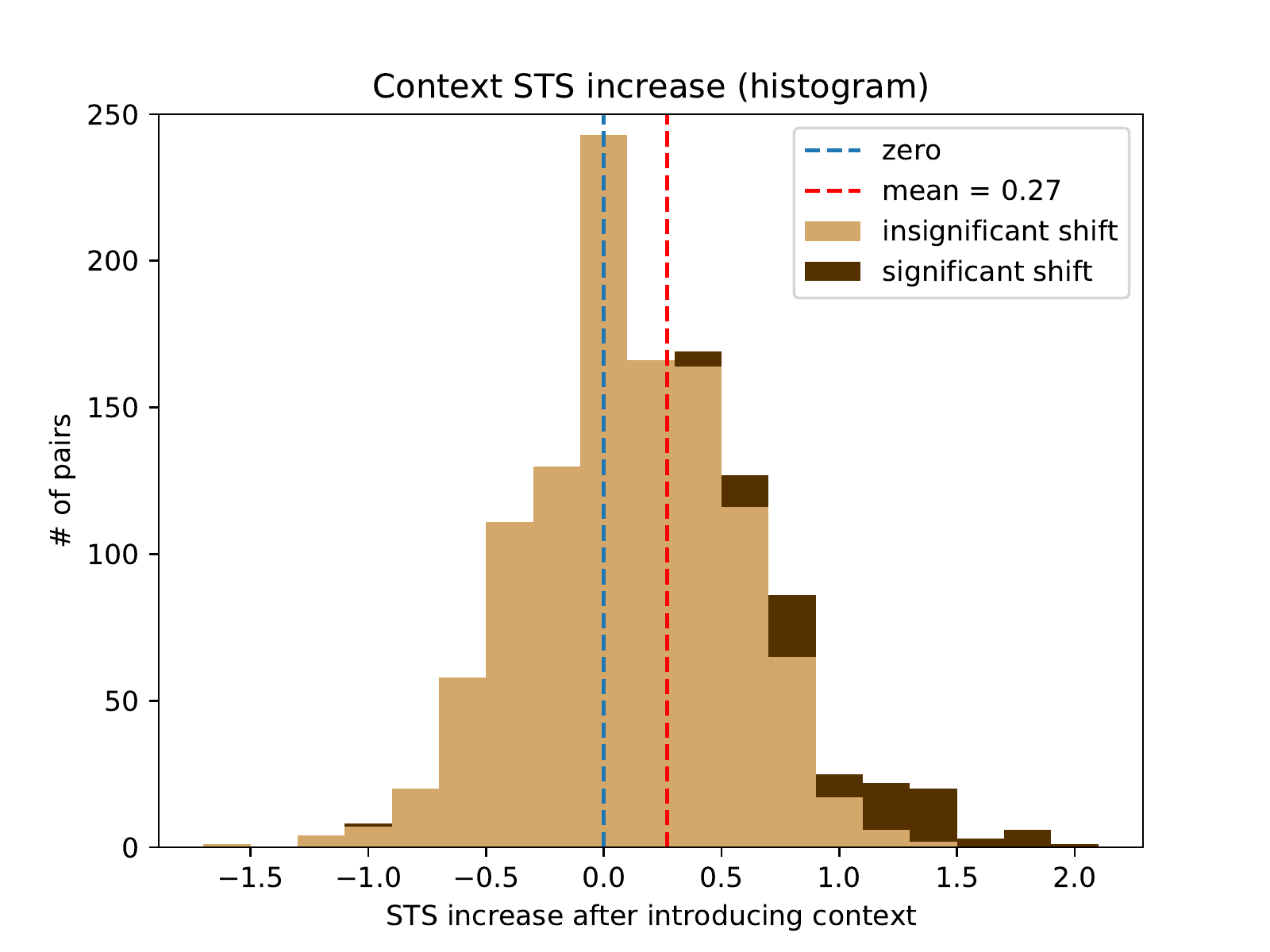}
        \caption{STS increase histogram (after context is introduced) of the whole test dataset.}
        \label{fig:sts_gain_1}
    \end{subfigure}
    \begin{subfigure}[b]{0.45\textwidth}
        \centering
        \includegraphics[width=\textwidth]{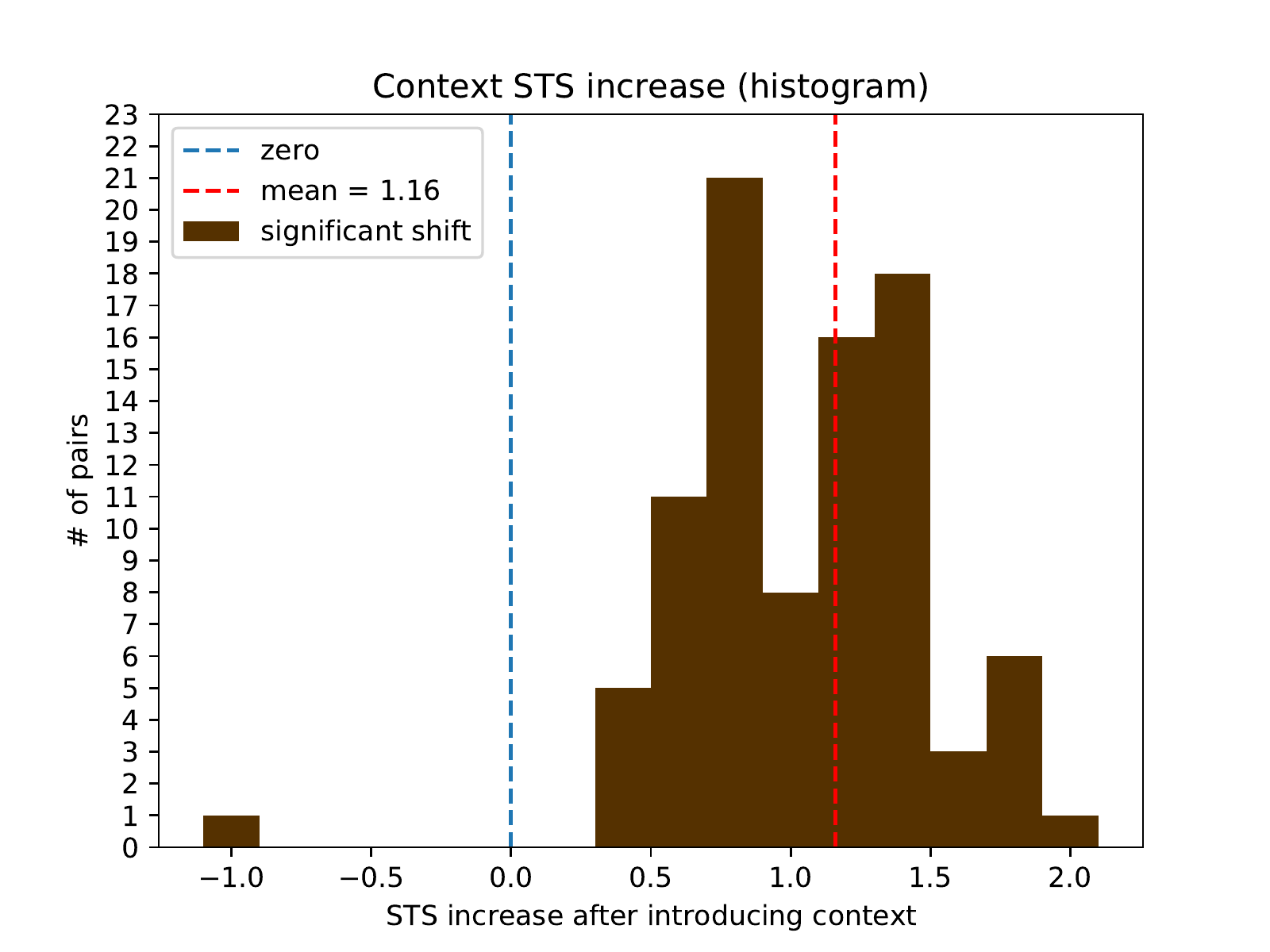}
        \caption{STS increase histogram of the significantly different ($\alpha=0.05$) pairs only.}
    \end{subfigure}
    \begin{subfigure}[b]{0.45\textwidth}
        \centering
        \includegraphics[width=\textwidth]{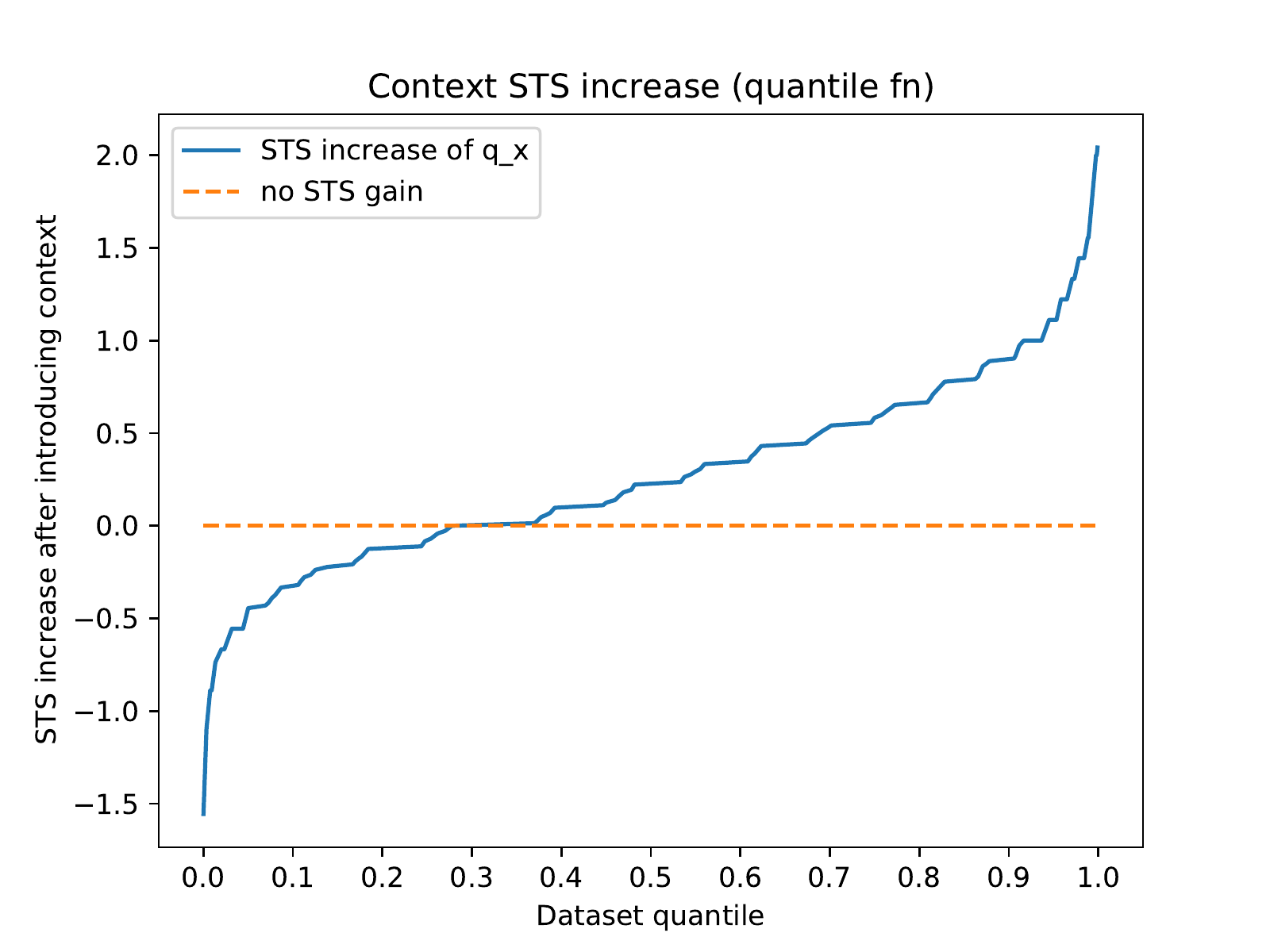}
        \caption{STS increase quantile function.}
        \label{fig:sts_gain_2}
    \end{subfigure}
    \caption{STS increase after introducing context to annotators in R2.}
        \label{fig:sts_gain}

\end{figure}

Figure \ref{fig:sts_gain} shows how much an STS score of a pair of sentences increases or decreases after the context is introduced to the annotators in the second round (R2). In Figure \ref{fig:sts_gain_1}, we can see that the mean of the increase is greater than zero (the exact difference is +0.26 of the STS score). The original distribution does not pass any normality test (because of its leptokurticity). However, if we think of the STS gains around 1.5 as outliers, it is safe to consider the distribution of gains to be approximately normal with $\mu = 0.268$ and $\sigma = 0.504$. The apparent steps in Figure \ref{fig:sts_gain_2} are caused by the non-continuous distribution of means of 9 whole numbers between 0 and 6.

\section{Initial Experiments}

To set up a baseline for the new dataset, we use the well-known and robust word2vec \cite{mikolov2013efficient} baseline and modern models based on the BERT architecture. Since the presented dataset contains data in Czech, we choose models pre-trained on the Czech language Czert \cite{sido2021czert} and SlavicBert \cite{arkhipov2019tuning-SlavicBert}. The BERT-based models can process sentences in two ways: a Cross-attention encoder on both sentences at once; and a Two Tower/Siamese encoder for each sentence separately with a similarity measure on the top.

\paragraph{The Word2Vec model}
We use an \textit{unweighted average of word2vec embeddings} to encode  sentences. We compute the similarity score by applying the cosine similarity on the resulting sentence vectors.

\paragraph{Two Tower Model}
We employ a stack based on the \textit{pooler outputs} from Bert-like models to encode each sentence independently. Then we add the cosine similarity measure on top and tune the model on the training dataset. 


\paragraph{Cross-Attention Model}
For the cross-attention encoder, we use the pooler output with a projection layer of the size of 200 with RELU activation on top followed by one single neuron with linear activation to get the similarity measure.  Again, we tune the model on the training dataset. 

The results are shown in Table \ref{tab:results} and discussed in Section 9.



\begin{table*}[]
\begin{adjustbox}{width=\linewidth,center}
\begin{tabular}{llclcclcc}
\toprule
 &
   &
  W2V &
   &
  \multicolumn{2}{c}{Czert} &
   &
  \multicolumn{2}{c}{Slavic Bert} \\ \cline{3-3} \cline{5-6} \cline{8-9} 
 &
   &
  \multicolumn{1}{l}{} &
   &
  Cross Attention &
  Two Tower &
   &
  Cross Attention &
  Two Tower \\ \midrule
\begin{tabular}[c]{@{}c@{}} MSE \\
  Pearson\\ Spearman \end{tabular} &
   &
  \begin{tabular}[c]{@{}c@{}}1.5043 $\pm$ 0.01979\\
  82.8300 $\pm$ 0.0702\\ 73.8225 $\pm$ 0.0783\end{tabular} &
   &
  \begin{tabular}[c]{@{}c@{}}1.1181 $\pm$ 0.0418\\
  91.887 $\pm$ 0.1193\\ 89.291 $\pm$ 0.1675\end{tabular} &
  \begin{tabular}[c]{@{}c@{}}1.7756 $\pm$ 0.0188\\ 
  88.177 $\pm$ 0.02407\\ 85.568 $\pm$ 0.06162\end{tabular} &
   &
  \begin{tabular}[c]{@{}c@{}}1.3483 $\pm$ 0.0381\\
  91.383 $\pm$ 0.2914\\ 88.966 $\pm$ 0.0892\end{tabular} &
  \begin{tabular}[c]{@{}c@{}}2.0352 $\pm$ 0.0414\\
  86.158 $\pm$ 0.1573\\ 83.634 $\pm$ 0.1500\end{tabular}
  \\\bottomrule
\end{tabular}
\end{adjustbox}
\caption{We report MSE -- Mean Square Error (first row), Pearson (second row) and Spearman (third row) correlations. The correlation coefficients are multiplied by a factor of 100.}
\label{tab:results}
\end{table*}

\section{Dataset Filtering}
\label{sec:postprocessing}
During post-processing, we clean the training part of the dataset by statistical comparison with the testing part. We suppose that the test part contains reliable annotations due to the exploratory phase and a carefully designed second round. The final scores come from averaging nine numbers making the test part more robust. 

\begin{figure}[ht!]
        \centering
    \includegraphics[width=\linewidth]{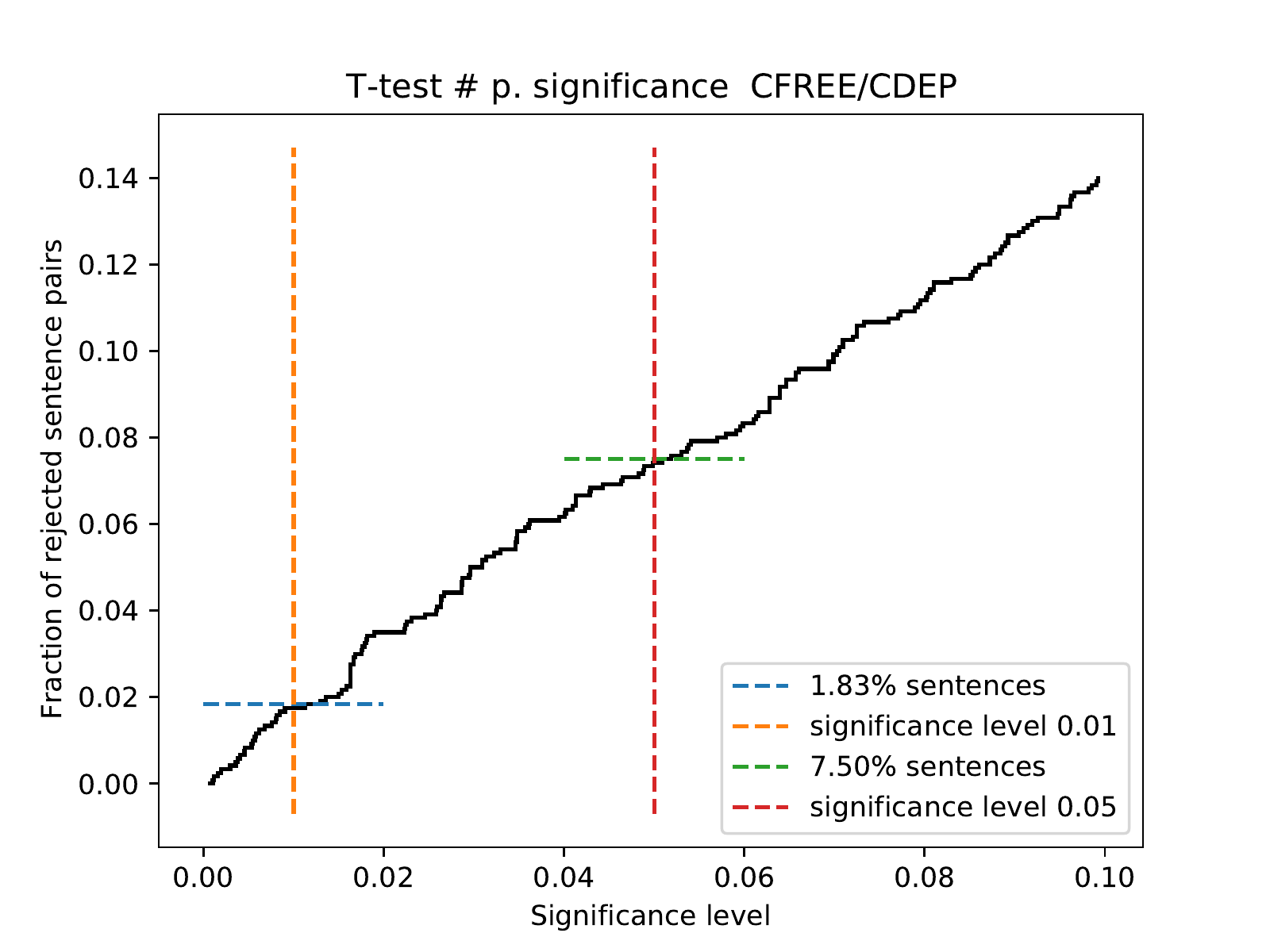}
    \caption{Evaluation of contextual semantic shift. The black line is a plot of $r(x)$. The dashed lines show the most common levels of significance. For the \textit{test-sig} version of the dataset, we cherry-picked samples showing change on significance level 0.05. } 
    \label{fig:semantic_drift}
\end{figure}
\paragraph{Filtering Regarding to the Semantic Shift}
\label{sec:Semantic_shift}
We design the process to collect context-free and context-dependent aligned data to enable future researchers to examine the role of context in semantics. 
We do not perform any special pre-filtering of the sentences presented to the annotators, so the natural (unbiased) distribution of contextual semantic shift should appear in the collected data.

We decided to quantify the significance of context by performing a series of {\it t}-tests, specifically to test the significant difference of means between the context-free and context-dependent main blocks element-wise (= one test for each sentence pair). We have assumed that the means of STS of size of nine samples for a single sentence pair and context presence are approximately normally distributed. 

To recapitulate -- for each sentence pair (of the main blocks), we possess nine STS scores for a~context-free and nine STS scores for a contextual version of the sentence pair. Then, we perform a~two-sided {\it t}-test for the equivalence of the corresponding means. The null hypothesis of this test is that the STS score means are equal; in other words, the added context is insignificant in the domain of STS. Each such test yield a certain {\it p}-value, that is, the largest possible significance level under which the null hypothesis is not rejected. The function $r(x)$ is defined as the fraction of sentence pairs for which we reject the null hypothesis of context-insignificance, if our level of significance is $x$, i.e., the CDF of the distribution of {\it p}-values (See Figure \ref{fig:semantic_drift}).

We can observe that the {\it p}-value of 0.05 yields 7.50 \% of sentences with different context annotations and the {\it p}-value of 0.01 results in 1.83 \%. We derive a new dataset from the sentences with significantly different context annotations at the {\it p}-value of 0.05. We name these dataset variants the \emph{test-sig free} and the \emph{test-sig dep} for context free and context dependent annotation respectively. Table \ref{tab:dataset_stats} shows additional statistics (annotation score mean, variance, count) of the datasets.

\begin{figure}[h!]
    \centering
     \centering
    \includegraphics[width=\linewidth]{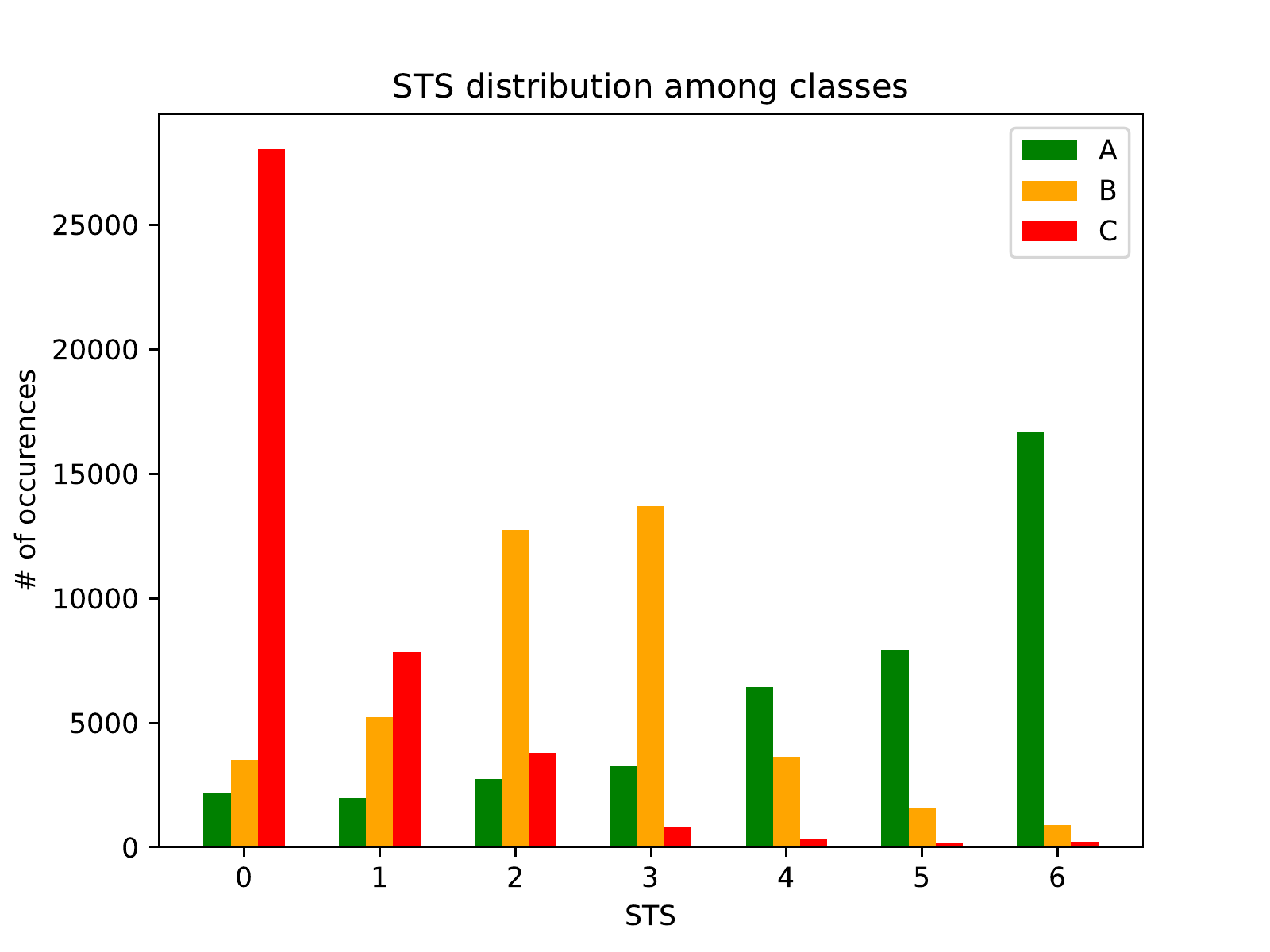}
    \caption{Motivation for diagonal filtering: Analysis of the frequency of final semantic similarity numbers grouped by suggested classes in the first round (R1). }
    \label{fig:diagonal_distribution}
  
\end{figure}

\paragraph{Diagonal Filtering of Training Dataset}
\label{sec:diagonal_filtering}
After the annotations, we compute some extra statistics of the collected data. We found out that there were some borderline cases in the first phase.  We made a frequency analysis of each rating (0--6) for three suggested classes (green, orange, red) \obrazek{fig:diagonal_distribution}. Due to the expected outcome, we were surprised by samples annotated against this scheme. There was small, but non-zero number of samples, picked for \textit{green} (should be similar) but rated with low numbers of similarity and wise versa, red ones (should not be similar) rated with high numbers of similarity.

We suppose this is possible due to several reasons. If we omit human mistakes, the most probable is that the data displayed to the annotator was not possible to annotate differently. There could be only a minor difference between reports and summaries so that the annotators could find no strongly dissimilar sentence, and they were forced to pick a~similar one even for the red class. However, we believe that such non-trivial examples were presented to annotators only in the minority. 

There are not so many possible reasons for data being biased oppositely. Again, if we omit systematic human mistake, which is unlikely, the only reason for such a systematic bias (observable also on orange suggested class) -- the annotators had no other choice. 
The only reasonable source of such bias is the presence of totally new information in the summary or sentences unused in original reports. We checked the possibility with the journalist -- They call it backgrounds, and the reason for using it is to place the summary into some (typically prior) context. And, of course, they often rephrase original sentences and underlay original pieces of information while creating a summary. 

Collecting the data in the first round could potentially bring systematic bias into training data -- humans marking different pairs as green tents to select higher similarity numbers. 
We decided to investigate this hypothesis by filtering of possible systematic bias in the training dataset by using only the close neighborhood of 1 and 2 score difference of intended colored class ($N1, N2$), where the number is the maximal difference of score from the expected value (green=6, orange=3, red=0). For statistics and results of basic experiments, see the Tables \ref{tab:corr}, \ref{tab:diagonal} and Figure \ref{fig:sts_distributions}.

Table \ref{tab:corr} also shows how significant an impact our R1 dataset filtering methods had on the correlation of STS between the corresponding R1 and test datasets. We can see a slight improvement in correlation after filtering using \textit{N2}, and yet another tiny improvement after filtering using \textit{N1}. To prove the correlation improvement is not caused by shrinking the intersection size, we sample a cropped test dataset, calculate the correlation between it and the corresponding part of the full R1 dataset, and average the results. The notable difference is between \textit{N1} and \textit{N1-random}, which shows that it is not sufficient to simply remove random 287 elements from the test dataset to improve the correlation.

\begin{table}[ht!]
\centering
\begin{tabular}{cccc}
\toprule
dataset         & mean & variance/MSE & size \\\midrule
train-raw       & 2.46  & 2.12  & 116 956          \\
train-N2        & 2.56  & 2.15  & 101 413          \\
train-N1        & 2.53  & 2.26  & 85 374          \\
test free       & 2.66  & 1.78  & 1200          \\
test dep        & 2.92  & 1.65  & 1200         \\
test-sig free   & 1.41  & 2.02  & 90        \\
test-sig dep    & 2.57  & 1.85  & 90       \\
\bottomrule
\end{tabular}
\caption{Statistical indicators of the created datasets.}
\label{tab:dataset_stats}
\end{table}

The additional statistics (mean, variance and size) of the dataset variants are available in table \ref{tab:dataset_stats}.


\begin{table}[]
\begin{adjustbox}{width=\linewidth,center}
\begin{tabular}{cccc}
\toprule
                    & RAW & N2     & N1    \\ \midrule
size & 116956 (100 \%)     & 101413 (86.71 \%) & 85374 (73.00 \%) \\ \midrule
       
Czert-CA &
  \begin{tabular}[c]{@{}c@{}}
  91.887 $\pm$ 0.1193\\ 89.282 $\pm$ 0.1755 \end{tabular} &
  \begin{tabular}[c]{@{}c@{}}
  91.525 $\pm$ 0.2343\\ 89.346 $\pm$ 0.1906\end{tabular} &
  \begin{tabular}[c]{@{}c@{}}
  91.25 $\pm$ 0.1812\\ 89.10 $\pm$ 0.09493\end{tabular} \\ \midrule
Pavlov-CA &
  \begin{tabular}[c]{@{}c@{}}
  91.383 $\pm$ 0.2914\\ 88.966 $\pm$ 0.0892\end{tabular} &
  \begin{tabular}[c]{@{}c@{}}
  91.14 $\pm$ 0.2638\\ 89.056 $\pm$ 0.1036\end{tabular} &
  \begin{tabular}[c]{@{}c@{}}
  91.039 $\pm$ 0.3166\\ 89.034 $\pm$ 0.1087\end{tabular}
  \\ \bottomrule
\end{tabular}
\end{adjustbox}
\caption{Filtering experiments. The size is shown absolute and relative numbers (the relative numbers are in brackets). We report Pearson (first line) and Spearman (second line) correlations multiplied by a~factor of 100. CA stands for cross-attention model.}
\label{tab:diagonal}
\end{table}

\section{Data Format}
Due to reasons described in Section \ref{sec:dataset}, we can release the dataset containing sentences with a limited surrounding context. 
We perform the initial experiments only with those files. Therefore, future researchers can use the same data as we did. Unfortunately, we can not release the original database with raw data collected during the annotation process. 
We try to bring to the reader the best insight into the whole process, the original data character, and the quality of the new final annotated corpora.
We publish all versions with and without the performed filtering.

We present the collected data in textual files available on our website\footnote{https://air.kiv.zcu.cz/datasets/sts-ctk} and on github\footnote{https://github.com/kiv-air/Czech-News-Dataset-for-Semanic-Textual-Similarity}. 
The averaged numbers of annotations for the same pair and enumeration of all annotations are presented in the test part files.

We also present the test dataset filtered by the significance of change between context-free and context-dependent annotations labeled as \textit{test-sig.tsv}. The data samples are filtered on a 0.95 confidence level of being significantly different between context-free and context-dependent annotations in this file. 


				

Context-free and context-dependent test parts are naturally aligned. However, context-free is about 6 \% bigger than context-dependent.

The train part of the dataset consists of two sentences with the user's annotation made in R1. The test part contains two sentences followed by the averaged STS value from R2, all original annotations collected in R2, and the value from R1.

The key sentence is surrounded with {\tt<sent></sent>} marks in the context-dependent variants of the dataset.


\section{Discussion}

\paragraph{The Role of the Context} The main goal was to create a new Czech dataset for semantic textual similarity with the aligned context-free and context-dependent annotations and evaluate the importance of context in usual texts. 
As shown, the context significantly influences a subset of the collected data.  
In the narrow domain of news and their summaries, we observe 90 samples from 1200 (7.5~\%) to be significantly shifted. We gather these samples into separate datasets so future researchers can utilize modern context-aware models to show their benefits.


\paragraph{The Role of Diagonal Filtering}
Our initial motivation to filter the training data came from contradictory annotations collected in the first round. Measuring correlations between R1 and R2 annotations validates our suspicion -- correlation increases with a throwaway of contradictory samples. Such filtering could help some methods; however, we confirm a generally known paradigm by higher evaluation metrics on unfiltered versions -- deep models can benefit from noisy but larger datasets more than from the cleaner and smaller variants. Nonetheless, we decided to share both filtered datasets (N2, N1) publicly.

\paragraph{Means and variances} We evaluate the distribution of scores in dataset splits and their different versions. The statistics are summarized in Table \ref{tab:dataset_stats}. We can see that mean of the context-dependent test dataset is significantly higher than the mean of the context-free dataset. Such an observation does make sense because with a larger context more information is present in the text, and there is a higher chance for some thematic overlap. Next, the test dataset has a higher mean and lower variance. This is caused by averaging nine annotators' scores in the test dataset. According to central limit theorem the variance of the average is lower and the mean is closer to the center of the interval.   

\paragraph{Initial Experiments} The initial experiments indicate that the  state-of-the-art vanilla models outperform a random human with relative ease. A random human annotator reaches 0.832 of Pearson and 0.777 of Spearman correlation coefficients with the dataset test set. However, the cross-attention state-of-the-art model beats a random human with 0.9189 of Pearson and 0.8929 of Spearman correlations. We believe that such a big difference comes from the ability of a computer model to capture a consensus of a large human group. We can measure a performance higher than the performance of individual annotators since the test part of the dataset is an average of 9 annotations. 


\section{Conclusion}

We conclude our paper with a summary of the distinct features of the introduced dataset. The large size of the dataset (138,556 annotated sentence pairs) allows robust training and evaluation of semantic models. The dataset belongs among the most extensive non-English training resources for learning the semantics of a language. 

The testing part of the dataset contains annotations based upon a consensus of nine annotators. Moreover, we performed a detailed analysis of the resulting annotations and filtered out the unreliable ones. We compute the theoretical lower bound of MSE to be approximately 0.1731. This number is considerably lower (better) than the performance of a random human annotator. Therefore, the testing part enables the evaluation of well-performing models.

Finally, we show that our dataset supports the training of well-performing models for semantic similarity of sentences. Our cross-attention model significantly outperforms an average human annotator.

We offer our dataset and the models for semantic similarity publicly accessible for research purposes.


\section*{Acknowledgement}
This work has been partly supported 
by the Technology Agency of the Czech Republic within the ETA Programme -- project TL02000288. Computational resources were supplied by the project "e-Infrastruktura CZ" (e-INFRA LM2018140) provided within the program Projects of Large Research, Development and Innovations Infrastructures.


\bibliographystyle{acl_natbib}
{\small
\bibliography{cite}}


\end{document}